\documentclass[titlepage,11pt]{article}

\usepackage[myheadings]{fullpage}
\usepackage{pmetrika}
\usepackage{amsmath}
\usepackage{submit}
\usepackage{subcaption}
\usepackage{graphicx}
\usepackage{booktabs}
\usepackage[hidelinks]{hyperref}
\usepackage{caption}
\usepackage{subcaption}

\usepackage{longtable}

\begin{document}

\begin{titlepage}

\title{Clustering individuals based on multivariate EMA time-series data}



\linespacing{1}

\author{Mandani Ntekouli}
\author{Gerasimos Spanakis}
\affil{Department of Advanced Computing Sciences}
\affil{Maastricht University}

\author{Lourens Waldorp}
\affil{Department of Psychological Methods}
\affil{University of Amsterdam}
\author{Anne Roefs}
\affil{Faculty of Psychology and Neuroscience}
\affil{Maastricht University}
\vspace{\fill}\centerline{}\vspace{\fill}

\comment{This study is part of the project ``New Science of Mental Disorders" (\url{www.nsmd.eu}), supported by the Dutch Research Council and the Dutch Ministry of Education, Culture and Science (NWO gravitation grant number 024.004.016).
}
\linespacing{1}
\contact{Correspondence should be sent to Mandani Ntekouli, Department of Advanced Computing Sciences of Maastricht University, Netherlands. E-Mail: m.ntekouli@maastrichtuniversity.nl}

\end{titlepage}

\setcounter{page}{2}
\vspace*{2\baselineskip}

\RepeatTitle{Clustering individuals based on multivariate EMA time-series data}\vskip3pt

\linespacing{1.5}
\abstracthead
\begin{abstract}
In the field of psychopathology, Ecological Momentary Assessment (EMA) methodological advancements have offered new opportunities to collect time-intensive, repeated and intra-individual measurements. This way, a large amount of data has become available, providing the means for further exploring mental disorders. Consequently, advanced machine learning (ML) methods are needed to understand data characteristics and uncover hidden and meaningful relationships regarding the underlying complex psychological processes. Among other uses, ML facilitates the identification of similar patterns in data of different individuals through clustering. This paper focuses on clustering multivariate time-series (MTS) data of individuals into several groups. Since clustering is an unsupervised problem, it is challenging to assess whether the resulting grouping is successful. Thus, we investigate different clustering methods based on different distance measures and assess them for the stability and quality of the derived clusters. These clustering steps are illustrated on a real-world EMA dataset, including 33 individuals and 15 variables. Through evaluation, the results of kernel-based clustering methods appear promising to identify meaningful groups in the data. So, efficient representations of EMA data play an important role in clustering.



\end{abstract}\vspace{\fill}\pagebreak


\section{Introduction}
In the course of EMA studies, time-intensive, repeated and intra-individual measurements are collected through digital questionnaires and smartphone’s app logs and sensors. Recent methodological advancements in collecting EMA data have offered new opportunities to collect a large amount of data on a personalized level, both in terms of time points and different variables of interest. Having more time points is always a desirable data characteristic, but when more variables are involved, training a linear Vector Autoregressive (VAR) model becomes computationally expensive, and sometimes even not feasible. 
Especially in a complex field as psychopathology, behaviors and psychological processes are prone to interact in a non-linear fashion. 
Thus, applying more complex and non-linear models becomes necessary.

Such complex models can be borrowed from the field of Machine Learning (ML). ML includes a wide range of advanced statistical and probabilistic techniques that learn to build models based on the provided data \cite{han2022data}. As a result, those models are able to uncover hidden characteristics and patterns in data. 
A popular example is through unsupervised clustering analysis. One application of clustering in EMA data can be to identify similar individuals \cite{genolini2016kmlshape}. Although all individuals exhibit their own characteristics, they may share common influences that lead to some similar behavior. So, information of people belonging to similar groups could potentially improve the baseline personalized models \cite{ntekouli2022using}.

This paper focuses on clustering multivariate time-series (MTS) data of different individuals into several groups. 
For clustering time-series data, various decisions should be made regarding the clustering algorithm, distance metric and the optimal number of clusters. Thus, the most efficient methods for these decisions are described in great detail. 
Finally, it is proposed that validation is performed through intrinsic methods examining quality and stability of clusters. This is an important part of this paper, given that validation of time-series clustering is considered as the most challenging part.






\section{Background on EMA time-series data characteristics}
Before describing the clustering process, an introduction to EMA time-series' characteristics is necessary. A key point, as well as a challenge of the current problem, is the multi-level structure of EMA data. During an EMA study, data are collected sequentially, at fixed time-intervals for all participating individuals.
An example could be every 2 hours for a period of 2-4 weeks. As a result, the captured data represent different aspects of participants' emotions over time and other contextual information.

When observing such a dataset, more special characteristics appear and need to be taken into account.
First, some measurements can be missing, mostly because of a machine or human error. This leads to datasets with incomplete time-series. 
Missing points affect also the time intervals between two consecutive measurements. When missing points exist, data are characterized as irregularly spaced MTS. 
In such cases, beyond deletion and imputation strategies, there are still ways to process data with missing values without relying on possibly biased techniques. A widely proposed approach is to apply a kernel to the raw data. Kernel methods have dominated ML because of their effectiveness in dealing with a variety of learning problems. To tackle these problems, a kernel can be applied to map data to a reproducing kernel Hilbert space (RKHS), that is higher dimension feature space. The success of kernel methods relies on the fact that nonlinear data structures, like high dimensional MTS, can be transformed based on the type of kernel to a space where they are finally linearly separable. 



Apart from length invariances, resulting from missing values, EMA time-series data can also exhibit different characteristics in terms of measurement scale and shift invariances. Regarding scaling, although EMA responses are usually recorded on a Likert scale, where 5 or 7 categories are available, the range of given responses may differ per participant. For example, some individuals may tend to be biased towards the middle values, avoiding all the extreme scores, whereas others may do the opposite, resulting in a higher skewness in some items, like negative emotions.  
In such cases, data normalization or scaling is a useful approach, whose effect is shown in Figure \ref{fig:EMA2}. 

Additionally, different individuals' time-series can exhibit shift invariances. Time-series represent the evolution of individual’s emotion or behavior. Thus, among different individuals, similar patterns of a behavior can be seen shifted in time. To be able to identify these shifted patterns and consider them as similar, an appropriate alignment method should be applied. For instance, alignment issues can be taken into account by an appropriate distance measure such as DTW, that will be further discussed later.

Before applying clustering, all the aforementioned special characteristics of time-series should be taken into account \cite{paparrizos2015k}. Thus, preprocessing and efficient data representations are required as additional steps. 


\section{Clustering Methodological Steps}
In this section, an overview of all the necessary steps and decisions for applying an EMA clustering is given. 
We examine all the decisions regarding distance metrics and clustering methods as well as how clustering options and results can be efficiently evaluated \cite{von2010clustering}. 

\subsection{Distance Metric}\label{dist}
Clustering algorithms are always relying on finding the most similar elements of a dataset and group them together. Similarity can be estimated by various distance metrics, each one reflecting a different characteristic, such as intensity or shape. In order to pick an adequate distance measure, the data variances, described before, have to be considered, otherwise, different clustering methods applied on the same dataset, can produce different results.

The most commonly used distance metric is the Euclidean distance, which can be used for both, tabular data and time series. A necessary requirement is that the different time-series should be of the same length. However, in the case of EMA datasets, this requirement is usually not satisfied because of missing values. A difference in the amount of missing values occurring in the data representing various individuals make the MTS to be of variant lengths.

\begin{figure}[t]
    \centering
    \subcaptionbox{\label{fig:EMA1}}
    {
        \includegraphics[width = 0.4\textwidth]{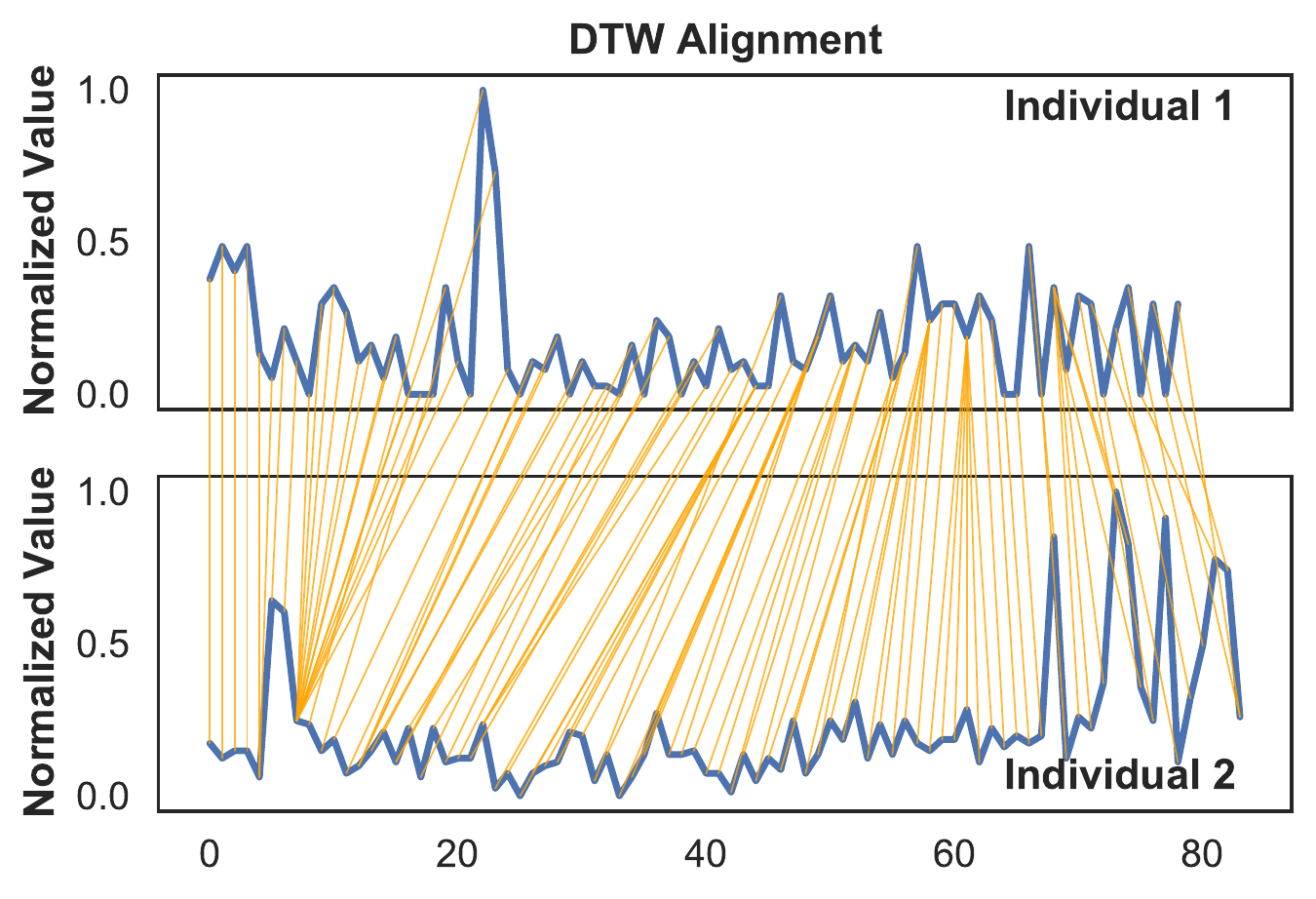}
    }
    \subcaptionbox{\label{fig:EMA2}}
    {
        \includegraphics[width = 0.4\textwidth]{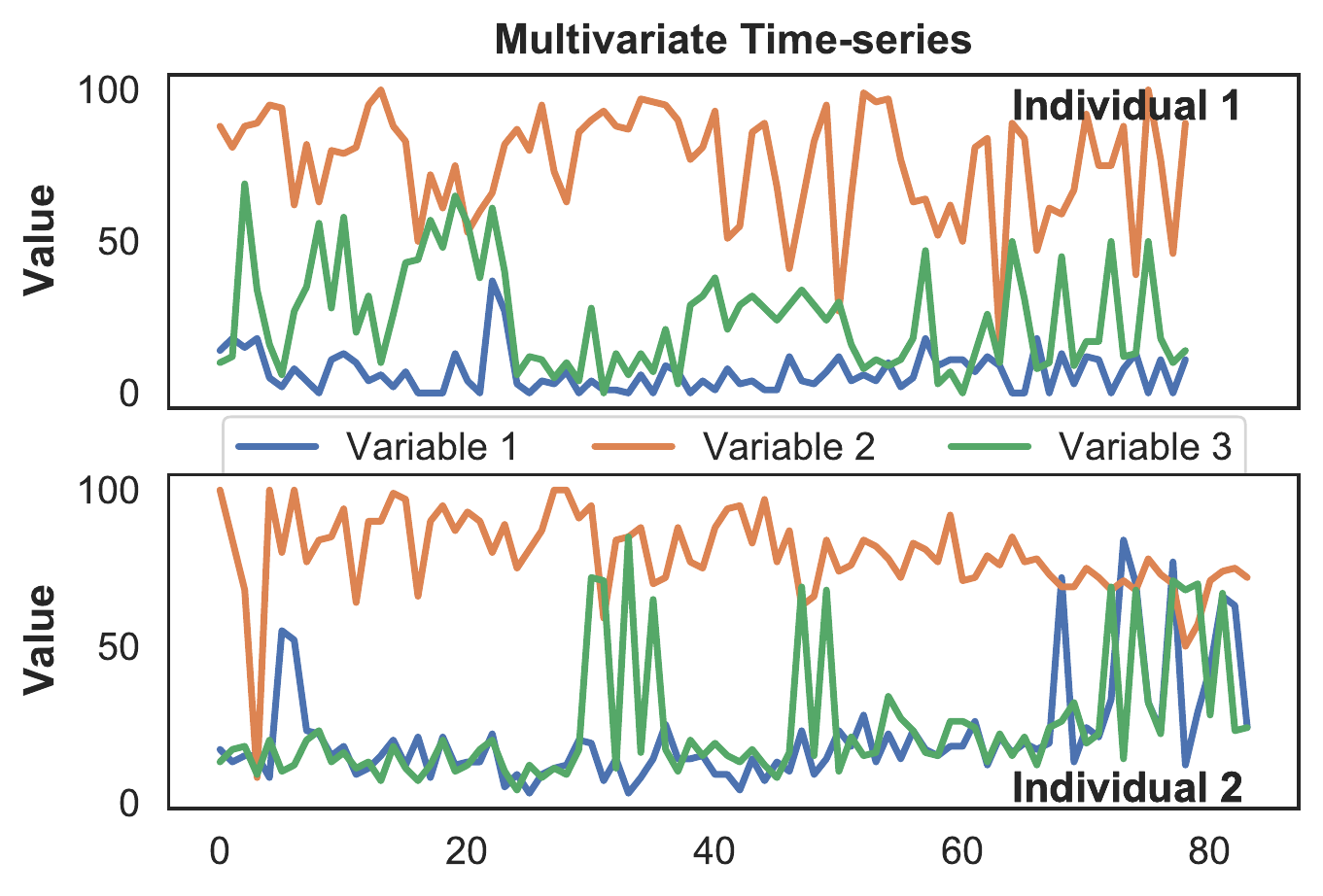}
    }
    \caption{
        \textbf{(\ref{fig:EMA1})} An example of 3 variables over time for 2 different individuals. \textbf{(\ref{fig:EMA2})} Best alignment between 2 individuals according to all variables. For the illustration, only Variable 1 is shown.
    }
\end{figure}

To tackle this issue, another widely-used distance metric is Dynamic Time Warping (DTW) which has become the state-of-the-art metric because of its high accuracy and its application in case of variable-length time-series \cite{sakoe1978dynamic}, \cite{javed2020benchmark}. Compared to Euclidean distance, DTW takes into account the shape difference of time-series. By stretching or compressing time series along the time axis, DTW aims to find the best shape-based alignment of these. This way, it also accounts for differences in points’ time interval due to missing values, but at the same time, outliers or noise do not significantly affect it. In practice, this is possible by comparing all possible alignment paths and finally get the one leading to the minimum distance. 
An example of the best alignment between the same EMA item of two individuals is illustrated in Figure \ref{fig:EMA2}. The vertical lines indicate the best alignment, showing that the two time series may not be ``warped” one by one.

Any distance metric can be viewed as a kernel as long as it is also positive definite \cite{cuturi2011fast}. Due to DTW's success, it was first considered as a good candidate for a kernel, however that's not directly possible, since it is based on Euclidean distance, which does not satisfy all the properties of a positive definite kernel (conditional positive definite). 
Hence, an alternative version for a time-series kernel is discovered which is called global alignment kernel (GAK) \cite{cuturi2011fast}. More specifically, as GAK was based on softDTW \cite{cuturi2017soft}, it takes advantage of the distance score values found across all possible alignment paths, rather than the optimal path found by DTW. According to this perspective, two time-series are considered similar not only if they have at least one alignment with high score, but quite more efficient alignment paths.


\subsection{Clustering Methods}
Due to the heterogeneity of clustering methods, this paper is limited to representative-based algorithms. These are distance-based methods whose goal is to retrieve a number of clusters defined by some representative elements or objects, named cluster centers. Clustering methods can be divided into two main categories, hard and fuzzy clustering \cite{aghabozorgi2015time}, \cite{javed2020benchmark}, \cite{ozkocc2020clustering}. In hard clustering methods, such as k-means and hierarchical clustering (HC), each individual is assigned to one cluster based on the highest similarity to clusters' center. Two challenges arise: how to integrate the appropriate distance metric and how to calculate the centroid of a cluster in case it is needed.

Nevertheless, from a theoretical point of view, in the field of psychopathology, a hard clustering algorithm could not always be the most appropriate choice. Knowing that psychopathology is a dynamically evolving, rather than a fixed, health condition, makes the approach of allowing individuals belonging to different clusters a more realistic scenario. Since clusters can capture dynamics in different time periods, individuals might be better represented by more than one cluster.
Furthermore, the fact that comorbidities, meaning the co-occurrence of many mental disorders, is prevalent in a high degree leads to shared psychological processes or behaviors among patients with different diagnoses \cite{roefs2022new}. Thus, clustering algorithms permitting individuals not to be strictly assigned to only one group are considered more plausible. This can be achieved by applying fuzzy clustering algorithms, such as Fuzzy c-means (FCM) and Fuzzy k-medoids (FKM). 


\subsection{Clustering Evaluation}
A ``good" clustering result is one that identifies the ``optimal" number of clusters and also how good objects, or individuals in this case, are grouped into clusters. Investigating how ``good" a clustering result is can be quite challenging, since usually, there are no ground truth labels (as in supervised tasks) to compare against. To overcome this issue, an intrinsic evaluation is performed. 


Ad-hoc intrinsic evaluation methods assign scores to a clustering result based on cohesion and separation. 
Some popular methods are Inertia, Silhouette Coefficient and Davies-Bouldin Index \cite{han2022data}. Out of these, Silhouette coefficient is picked as a metric, since it takes into account both intra-cluster and inter-cluster similarities. It compares the average similarity across individuals of the same cluster to the points belonging to the closest one. To find the closest cluster, similarities among all individuals in a cluster is taken into account. Thus, it’s quite straightforward to interpret the clustering results. Its values range from -1 to 1, where 1 and -1 indicate the best and the worst clustering, respectively, whereas 0 show a meaningless grouping, for example, when similarity differences between clusters are negligible. On the other hand, in case of fuzzy clustering, additional evaluation measures have been widely adopted, further assessing the membership degree of each individual into different groups \cite{choudhry2016performance}. The most common ones are Partition Coefficient (PC), Partition Entropy (PE) and Xie-Beni (XB) index, all examining the fuzziness of individuals in a different way. Apart from PC (ranges from 0 to 1), PE and XB are not bounded, while the optimal number of clusters is found at the highest, lowest and highest values, respectively.  
Consequently, these estimates give more information about the efficiency of fuzzy clustering.

Moreover, the stability of the clustering result should be taken into account. Running a clustering algorithm multiple times may lead to different results due to different initialization values. 
To evaluate clustering stability, it is needed to run the clustering algorithm several times and compare the matching of individuals’ cluster assignment. After checking all label permutations, the produced distance quantifies the mean cluster disagreement across all pairs of individuals. The result represents the clustering instability index (called stability by \cite{von2010clustering}) and its value can range from 0 (most stable) to 1 (less stable). 

Furthermore, the extracted evaluation coefficients (such as Silhouette) can also be tested for their consistency by investigating their distribution across different runs of the algorithm. If the coefficients vary a lot, then that is indication of an unstable clustering.

Summarizing, there are various methods for evaluating a good clustering approach. Thus, in this paper, a good clustering is defined as a combination of some of the aforementioned methods. More specifically, the number $k$ of clusters is primarily determined based on a high Silhouette coefficient, but this decision should be consistent to the findings of the other evaluation indexes as well. Subsequently, cluster stability requirement should also be fulfilled. Stability is examined on the instability index as well as the consistency of silhouette coefficients when clustering is repeatedly applied.

\section{Experimental Results}
In this section, an example real-world dataset is used to illustrate all the decisions about methods, presented in the previous sections. The used dataset is a real-world dataset obtained by a study described in \cite{soyster2022pooled}. It is a result of a 2-week data collection
from 33 individuals, providing roughly 89 data points per individual. In a goal to capture alcohol consumption, 15 variables/indicators (such as positive and negative emotions, drinking craving and expectancies) were included in the data collection.
We perform clustering on the 33 individuals based on their 15-variable time-series, taking into account the specific issues discussed in the previous chapter.
Following, clustering results are evaluated through examining cluster quality and stability. 




First, we apply clustering through 
k-means ($km_{DTW}$, $km_{GAK}$), HC ($HC_{DTW}$, $HC_{GAK}$) and FKM ($FKM_{DTW}$, $FKM_{GAK}$). 
Both distance metrics (DTW and GAK) are examined, except for fuzzy c-means (FCM) where only the DTW was used, as it is quite difficult to extract the clusters' centroids in the original dimensions, due to kernalization. The $\sigma$ hyperparameter of the GAK kernel depends on the given data and it is calculated as the average of the median of all distances \cite{cuturi2011fast}. Then, the groups derived by all clustering methods are evaluated in terms of Silhouette coefficient as well as stability. According to this, the optimal number of clusters is determined as well as the quality of the retrieved clusters.

\begin{figure}[t]
    \centering
    \subcaptionbox{\label{fig:sil_eval}}
    {
        \includegraphics[width = 0.40\textwidth]{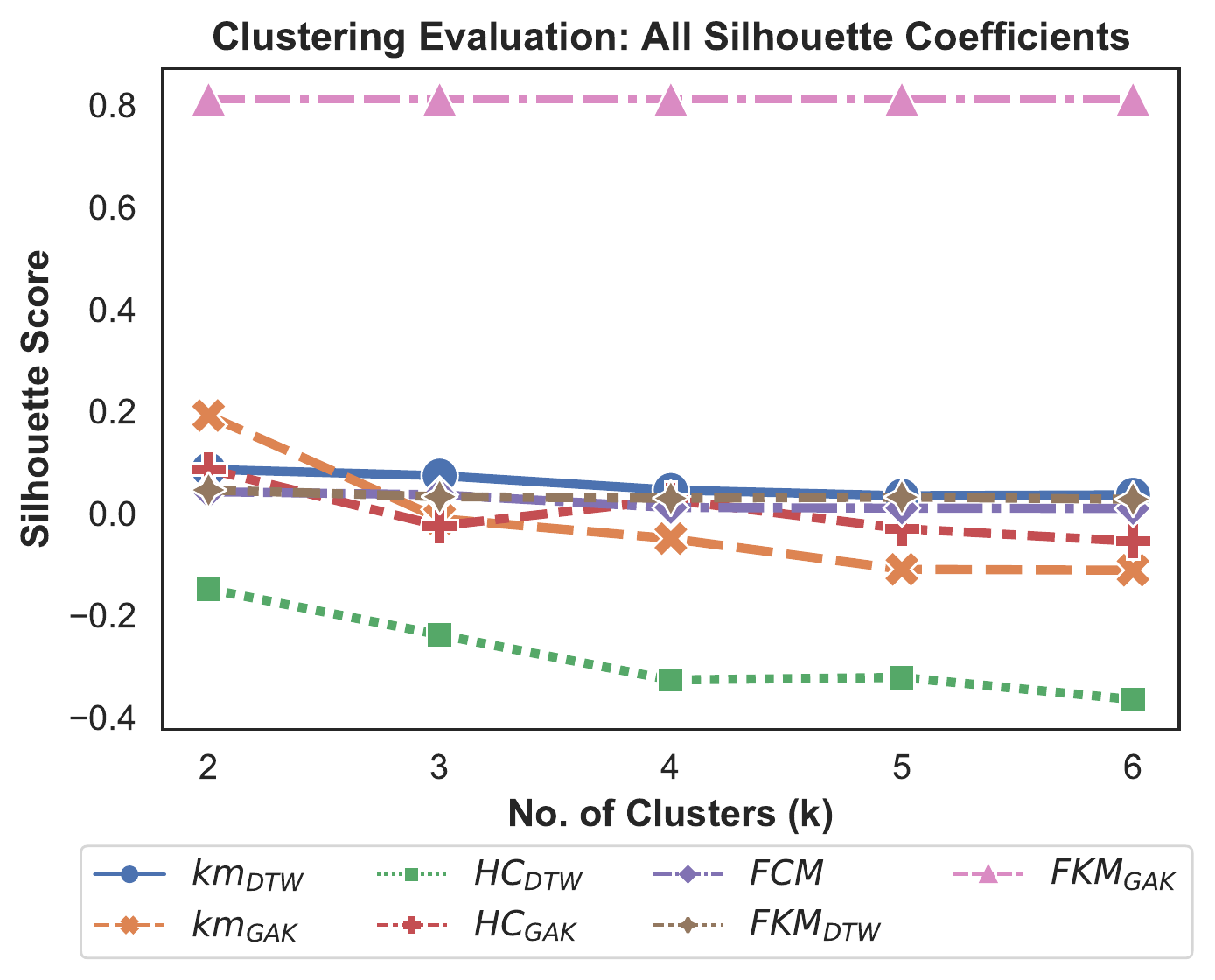}
    }
    \subcaptionbox{\label{fig:fuz_eval}}
    {
        \includegraphics[width = 0.45\textwidth]{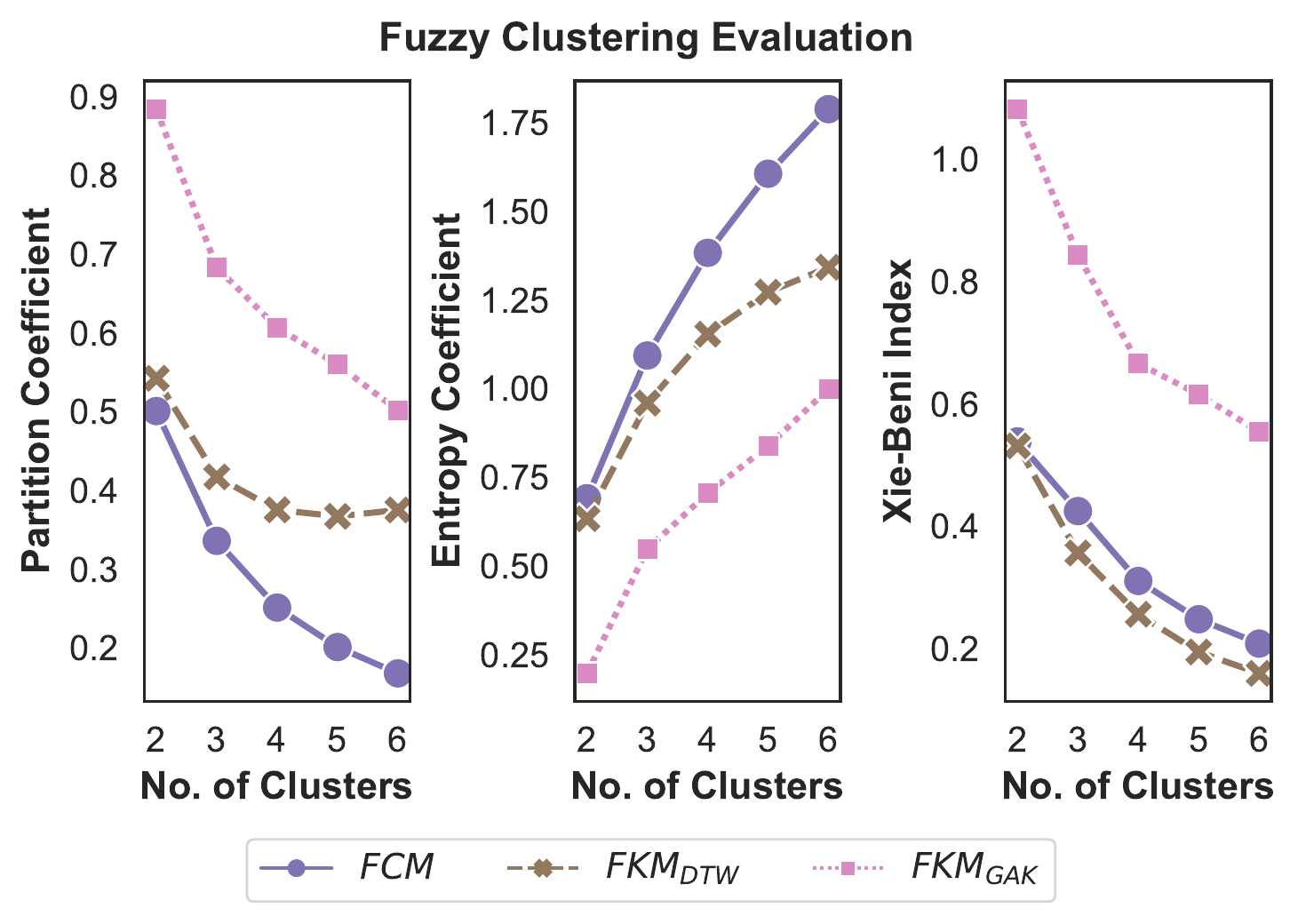}
    }
    \caption{
        \textbf{(\ref{fig:sil_eval})} Maximum Silhouette Scores for all algorithms. \textbf{(\ref{fig:sil_eval})} Overall clustering evaluation.
    }
\end{figure}

Regarding the Silhouette analysis, overall results are shown in Figure \ref{fig:sil_eval}.  We notice that the FKM method using a GAK kernel gives the highest score. It is interesting that this remains constant for different values of clusters, as they are always grouped to two clusters even in cases when more are allowed (leading to empty clusters). Also, a quite high score is produced by kernel k-means with $k=2$. Apart from these, the rest of the algorithms show a result close to zero, which is interpreted as a not so meaningful clustering result. The best result among these is given by HC using a GAK kernel with $k=2$. Therefore, it is interesting to observe that when a kernel-based method is utilized, the quality of the retrieved clusters seems to be better, showing that kernels are needed to better represent the complex structure of EMA data.

In case of fuzzy clustering methods, additional intrinsic evaluation measures can be used. The scores for different number of $k$ are presented in Figure \ref{fig:fuz_eval}. These appeared to be consistent to the Silhouette results, showing that $k=2$ is the optimal choice, also for the fuzzy clustering algorithms.

\begin{figure}[t]
    \centering
    \subcaptionbox{\label{fig:stab}}
    {
        \includegraphics[width = 0.35\textwidth]{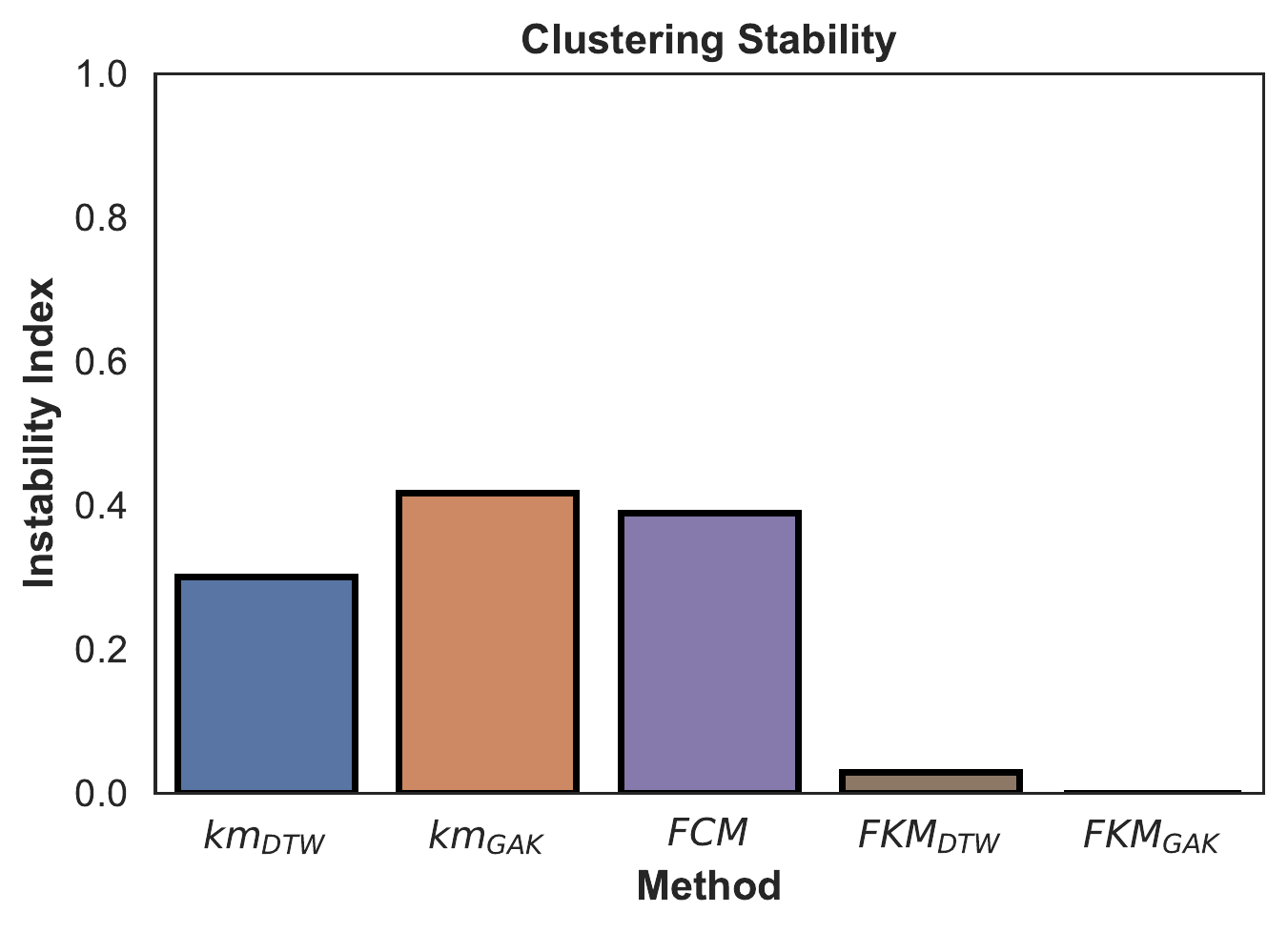}
    }
    \subcaptionbox{\label{fig:dist_sil}}
    {
        \includegraphics[width = 0.35\textwidth]{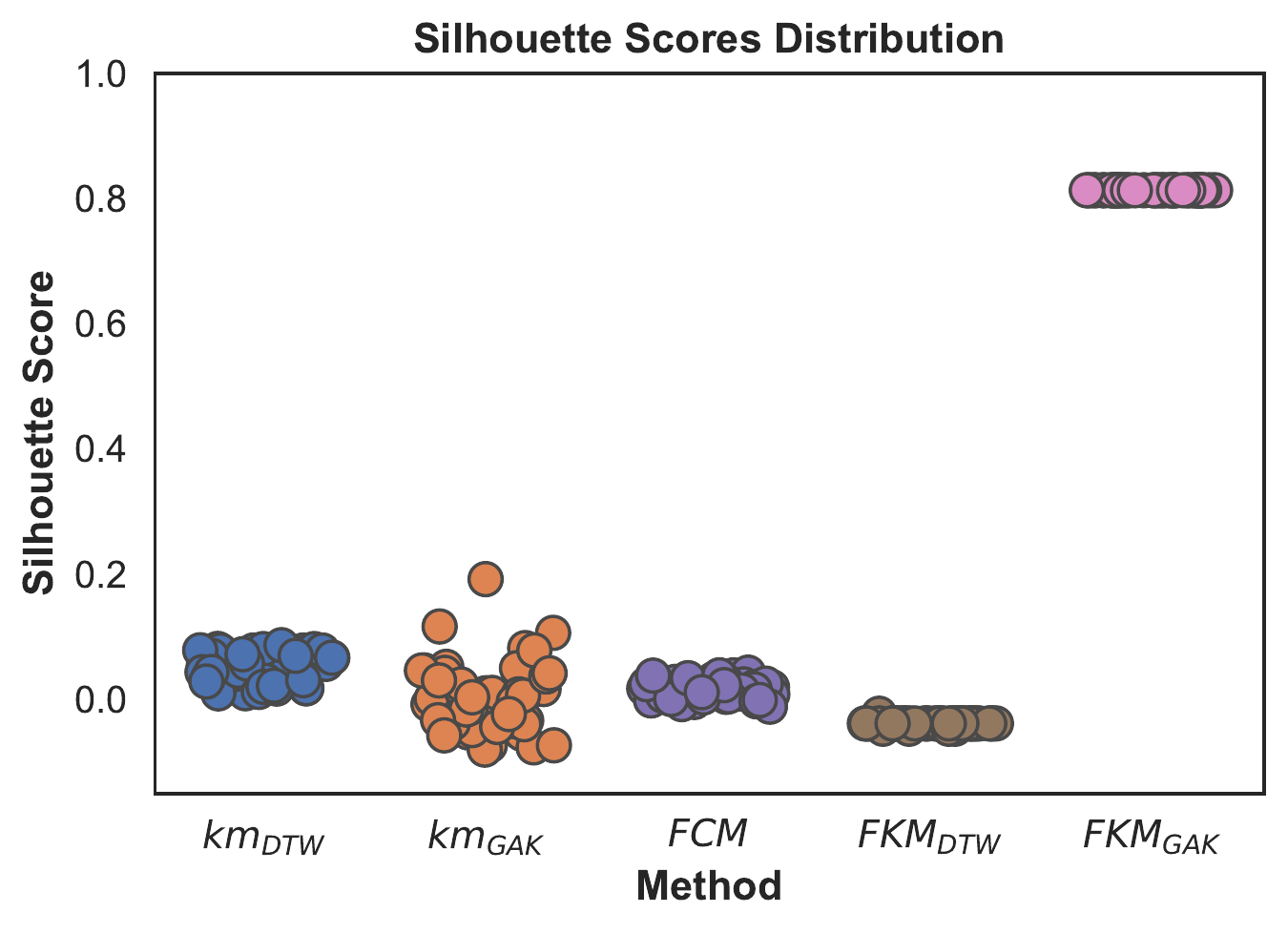}
    }
    \caption{
        Clustering evaluation for $k=2$. \textbf{(\ref{fig:stab})} Clustering instability index. \textbf{(\ref{fig:dist_sil})} Distributions of silhouette scores.
    }
\end{figure}

Next, we check the stability of the clustering-derived groups through silhouette scores consistency and instability index. 
Instability index and silhouette scores distribution were computed for 50 runs of each algorithm and are presented in Figures \ref{fig:stab} and \ref{fig:dist_sil}, respectively. For this part, HC is not included as it's independent of initialization issues. According to these figures, the most stable clustering result is produced by FKM, whereas the least stable by kernel k-means. A low instability score shows that groups' separation does not change a lot across repetitions. However, we can still observe an interesting case, or run, of an outlier in kernel k-means with a score approximating $0.2$, which is quite higher compared to the rest. This is also apparent in Figure \ref{fig:sil_eval}, for $km_{GAK}$ and $k=2$, and worth further investigating.

Summarizing, from a methodological perspective, various choices are possible for algorithms, distance metric and evaluation, which lead to different results. Although it is important that all methods extracted 2 clusters as the optimal grouping, it does not mean that individuals are assigned into groups in a similar way. This is also reflected when getting different results during evaluation. It is interesting to highlight that the method evaluated as the most stable is $FKM_{GAK}$, regardless of the issue of initial parameters. Also, the fact that always two clusters were retrieved, even though more were allowed, gave more evidence for the optimal number of clusters.


\section{Related Work}
As already discussed, applying clustering methods to time-series data has been widely explored. Some examples of review studies are \cite{aghabozorgi2015time}, \cite{javed2020benchmark}, \cite{ozkocc2020clustering}. Considering that all well-known clustering algorithms can be used for time-series, the challenge becomes on how to pick the right distance metric. Thus, most research studies have focused on finding a good representation of time-series similarities and integrate it to clustering algorithms. 

Due to the success of the shape-based time-series clustering, other DTW-variations have been suggested, by either applying some restrictions on DTW or softening the optimal distance paths using softDTW \cite{cuturi2017soft}. Other studies exploring different shape-based information (\cite{vlachos2002discovering},
\cite{paparrizos2015k} and \cite{genolini2016kmlshape}),
propose the use of the longest common subsequence (LCSS), cross-correlation and Fréchet distance, respectively.

However, most studies have handled univariate time-series data. The added value of the current paper is the multi-level structure of EMA data, including several multivariate time-series. In case of multivariate time-series, kernel-based data representations have been proposed \cite{badiane2018kernel}. Kernels based on DTW, such as GAK, were used \cite{cuturi2017soft}. Moreover, in \cite{mikalsen2018time}, another time-series cluster kernel (TCK) was proposed, based on Gaussian mixture models (GMMs).

Specifically for EMA data, only little research work has been conducted as far as clustering is concerned. In \cite{torous2018smartphones}, clustering EMA data into similar meaningful groups or clusters is proposed. However, it was not applied leaving a gap that is covered in this paper. Other than this, a different goal focusing on clustering EMA items was investigated in \cite{hebbrecht2020understanding}. In that case, clustering was used to organize a person’s symptomatology into homogeneous categories of symptoms and not for grouping different individuals like in the current paper.

\section{Conclusions}
This paper aims to address some of the challenges of EMA data modeling by grouping or clustering similar individuals. A detailed review of all the potential directions for applying clustering based on time-series patterns. Having described the heterogeneity of existing methods, the focus was then placed on the most challenging part of clustering, which is evaluation. A combination of several well-known ad-hoc evaluation measures was proposed, examining clustering quality through Silhouette coefficients as well as stability. According to our analysis, kernel-based clustering methods produced the best quality clusters, showing that kernels can be useful for efficient EMA data representations. Future work can include a simulation study for evaluating clustering methods in different EMA experimental scenarios as well as further exploration of data representations using different kernels, since it plays an important role in clustering. Moreover, it should be investigated how clustering-derived groups of individuals could be further utilized. For example, an interesting approach is to train group-based models for providing more accurate predictive capabilities.

\bibliographystyle{apalike} 
\bibliography{Refs.bib} 


\end{document}